\documentclass[graybox]{svmult}

\usepackage{mathptmx}       
\usepackage{helvet}         
\usepackage{courier}        
\usepackage{type1cm}        
\usepackage{makeidx}         
\usepackage{graphicx}        
\usepackage{multicol}        
\usepackage[bottom]{footmisc}

\usepackage{algorithm}
\usepackage{algorithmic}

\usepackage{amsmath,amsfonts,amssymb,mathrsfs}
\usepackage[fleqn,tbtags]{mathtools}

\newcommand{\beq}{\begin{equation}} 
\newcommand{\eeq}{\end{equation}}

\mathtoolsset{showonlyrefs,showmanualtags}

\newcommand{\la}{{\langle}}
\newcommand{\ra}{{\rangle}}

\newcommand{\reg}{{\omega}}

\newcommand{\bi}{\begin{itemize}}
\newcommand{\be}{\begin{enumerate}}
\newcommand{\ei}{\end{itemize}}
\newcommand{\ee}{\end{enumerate}}

\newcommand{\R}{{\mathbb R}}
\newcommand{\N}{{\mathbb N}}

\newcommand{\lam}{{\lambda}}

\newcommand{\lb}{{\langle}}
\newcommand{\rb}{{\rangle}}

\def\boldf#1{\hbox{\rlap{$#1$}\kern.4pt{$#1$}}}

\newcommand{\trans}{^{\scriptscriptstyle \top}}

\DeclareMathOperator\argmin{argmin}

\DeclareMathOperator\prox{prox}

\newcommand{\uh}{\hat{y}}

\begin{document}

\title*{On Sparsity Inducing Regularization Methods for Machine Learning}

\author{Andreas Argyriou, Luca Baldassarre, Charles A. Micchelli and Massimiliano Pontil}
\institute{Andreas Argyriou \at Ecole Centrale de Paris,
Grande Voie des Vignes 92 295 Chatenay-Malabry, FRANCE. \email{andreas.argyriou@ecp.fr} \and Luca Baldassarre \at Laboratory for Information and Inference Systems, EPFL, ELD 243, Station 11 CH-1015 Lausanne, Switzerland.  \email{luca.baldassarre@epfl.ch} \and Charles A. Micchelli \at Department of Mathematics and Statistics, University at Albany, Earth Science 110 Albany, NY 12222, USA. \email{charles\_micchelli@gotmail.com} \and Massimiliano Pontil \at Department of Computer Science, University College London, Malet Place
London WC1E 6BT, UK. \email{m.pontil@cs.ucl.ac.uk}}

%
%
\maketitle

{\em Dedicated to Vladimir Vapnik with esteem and gratitude for his fundamental contribution to Machine Learning.}

\vspace{1.5truecm}

\abstract{During the past years there has been an explosion of interest in learning methods based on sparsity regularization. In this paper, we discuss a general class of such methods, in which the regularizer can be expressed as the composition of a convex function $\omega$ with a linear function. This setting includes several methods such the group Lasso, the Fused Lasso, multi-task learning and many more. We present a general approach for solving regularization problems of this kind, under the assumption that the proximity operator of the function $\omega$ is available. Furthermore, we comment on the application of this approach to support vector machines, a technique pioneered by the groundbreaking work of Vladimir Vapnik.}

\section{Introduction}
In this paper, we address supervised learning methods which are based on
the optimization problem
\beq
\min_{x \in \R^d} \big\{f(x)+g(x) \big\},
\label{eq:opt}
\eeq
where the function $f$ measures the fit of a vector $x$ (linear predictor) to available
training data and $g$ is a penalty term or regularizer which
encourages certain types of solutions. Specifically, we let $f(x)=
E(y,Ax)$, where $E: \R^s \times \R^s \rightarrow [0,\infty)$ is an
error function, $y\in \R^s$ is a vector of measurements and $A \in \R^{s
\times d}$ a matrix, whose rows are the input vectors. This class of
regularization methods arise in machine learning, signal processing
and statistics and have a wide range of applications. 

 
Different choices of the error function and the penalty function
correspond to specific techniques. In this paper, we are interested in
solving problem \eqref{eq:opt} when $f$ is a {\em strongly smooth convex} function (such as the square error $E(y,Ax) =\|y-Ax\|_2^2$) and the penalty function
$g$ is obtained as the composition of a ``simple'' function with a
linear transformation $B$, that is,
\beq
g(x) = \omega(Bx) \; ,
\label{eq:comp-reg}
\eeq
where $B$ is a prescribed $m \times d$ matrix and $\omega$ is a 
{\em nondifferentiable convex} function on $\R^d$. 
The class of regularizers \eqref{eq:comp-reg}
includes a variety of methods, depending on the choice of the
function $\omega$ and of matrix $B$. Our motivation for studying
this class of penalty functions arises from sparsity-inducing
regularization methods which consider $\omega$ to be either the
$\ell_1$ norm or a mixed $\ell_1$-$\ell_p$ norm. When $B$ is the
identity matrix and $p=2$, the latter case corresponds to
the well-known Group Lasso method \cite{yuan}, for which well studied optimization
techniques are available. Other choices of the matrix $B$ give rise to
different kinds of Group Lasso with overlapping groups
\cite{Jenatton,binyu}, which have proved to be effective in modeling structured sparse regression problems. 
Further examples can be obtained by considering composition with the
$\ell_1$ norm, for example this includes the Fused Lasso penalty function
\cite{tib05} and the graph prediction problem of \cite{mark09}.

A common approach to solve many optimization problems of the 
general form \eqref{eq:opt}
is via proximal-gradient methods. These are first-order iterative methods, 
whose computational cost per iteration is comparable to gradient descent.
In some problems in which $g$ has a simple expression, proximal-gradient methods can be combined 
with acceleration techniques \cite{Nesterov83,Nesterov07,tseng10}, to 
yield significant gains in the
number of iterations required to reach a certain approximation 
accuracy of the minimal value. 
The essential step of proximal-gradient methods requires the computation of the proximity
operator of function $g$, see Definition
\ref{def:prox} below. In certain cases of practical importance,
this operator admits a closed form, which makes proximal-gradient methods
appealing to use. However, in the general case \eqref{eq:comp-reg}
the proximity operator may not be easily computable.

We describe a general technique to compute the proximity operator of the composite regularizer
\eqref{eq:comp-reg} from the solution of a fixed point problem, which depends on the proximity operator of the function $\omega$ and the matrix $B$. This problem can be solved by a simple and efficient iterative scheme when the proximity operator of $\omega$ has a closed form or can be computed in a finite number of steps. When $f$ is a strongly smooth function, the above result can be used together with Nesterov's accelerated method
\cite{Nesterov83,Nesterov07} to provide an efficient first-order method for solving the optimization problem \eqref{eq:opt}. 

The paper is organized as follows. In Section \ref{sec:2}, we review
the notion of proximity operator, useful facts from fixed point
theory and present a convergent algorithm for the solution of problem \eqref{eq:opt} when $f$ is quadratic function and then an algorithm to solve the associated optimization
problem~\eqref{eq:opt}. In Section \ref{sec:3}, we discuss some examples of composite
functions of the form \eqref{eq:comp-reg} which are valuable in
applications. In Section \ref{sec:4} we apply our observations to support vector machines and obtained new algorithms for the solution of this problem. Finally, Section \ref{sec:5} contains concluding remarks.



\section{Fixed Point Algorithms Based on Proximity Operators}
\label{sec:2}
In this section, we present an optimization approach which use fixed
point algorithms for nonsmooth problems of the form \eqref{eq:opt}
under the assumption \eqref{eq:comp-reg}. We first recall some notation and then move on 
to present an approach to compute the proximity operator for composite regularizers.

\subsection{Notation and Problem Formulation}

We denote by $\la\cdot,\cdot\ra$ the Euclidean inner product on $\R^d$
and let $\|\cdot\|_2$ be the induced norm. If $v:\R \rightarrow \R$, for
every $x \in \R^d$ we denote by $v(x)$ the vector $(v(x_i))_{i=1}^d$.
For every $p \geq 1$, we define the $\ell_p$ norm of $x$ as $\|x\|_p = 
(\sum_{i=1}^d |x_i|^p)^\frac{1}{p}$.

As the basic building block of our method, we consider the optimization problem \eqref{eq:opt}
in the special case when $f$ is a quadratic function and the regularization term $g$ is obtained 
by the composition of a convex function with a linear function. That is, we consider the problem
\beq
\min\left\{ \dfrac{1}{2}y\trans Q y - x\trans y + \reg(By) : y \in\R^d \right\} \,.
\label{eq:quad}
\eeq
where $x$ is a given vector in $\R^d$ and $Q$ a positive definite $d
\times d$ matrix. The development of a convergent method for the solution of this problem requires the well-known concepts of proximity operator and subdifferential of a convex function. Let us now review some of salient features of these important notions which are needed for the analysis of problem \eqref{eq:quad}. 

The proximity operator on a Hilbert space was introduced by Moreau in \cite{moreau62}. 

\begin{definition}
Let $\reg$ be a real valued convex function on $\R^d$. The proximity operator of $\reg$ is defined, for every $x\in\R^d$ by
\beq
\prox_\reg (x) := \argmin 
\left\{ \dfrac{1}{2} \|y-x\|_2^2 + \reg(y) : y \in \R^d\right\} \,.
\label{eq:prox}
\eeq
\label{def:prox}
\end{definition}
The proximity operator is well defined, because the
above minimum exists and is unique. 

Recall that the subdifferential of $\reg$ at $x$ is defined as
$\partial \reg(x) = \{u: u \in \R^d, \la y-x,u\ra +\reg(x) \leq \reg(y),~ \forall y \in \R^d\}$. The subdifferential is a nonempty compact and convex set. Moreover,
if $\reg$ is differentiable at $x$ then its subdifferential at $x$
consists only of the gradient of $\reg$ at $x$. 

The relationship between the proximity
operator and the subdifferential of $\reg$ are essential for algorithmic developments for the solution of 
\eqref{eq:quad}, \cite{andy-tech,combettes,MSX,mosci10}. Generally the proximity operator is difficult to compute since it is expressed as the minimum of a convex optimisation problem. However, the are some rare circumstances where it can obtained explicitly, for examples when $\omega(x)$ is a multiple of the $\ell_1$ norm of $x$ the proximity operator relates to soft thresholding and moreover a related formula allows for the explicit identification of the proximity operator for the $\ell_2$ norm, see, for example, \cite{andy-tech,combettes,MSX}. Our optimisation problem \eqref{eq:quad} can be reduced to the identification of the 
proximity operator for the composition function $\omega \circ B$. Although the prox of $\omega$ may be readily available, it may still be a computational challenge to obtain the prox of $\omega \circ B$. We consider this essential issue in the next section.


\subsection{Computation of a Generalized Proximity Operator with a Fixed Point Method}
\label{sec:quad}

In this section we consider circumstances in which the proximity operator of $\omega$ can be explicitly computed in a finite number of steps and seek an algorithm for the solution of the optimisation problem \eqref{eq:quad}.

As we shall see, the method proposed here applies for any positive definite matrix $Q$. This will allow us in a future publication to provide a second order method for solving \eqref{eq:opt}. For the moment, we are content in focusing on \eqref{eq:quad} by providing a technique for the evaluation of $\prox_{\omega \circ B}$.

First, we observe that the minimizer ${\hat y}$ of \eqref{eq:quad} exists and is {\em unique}. Indeed, this vector is characterised by the set inclusion
\beq
Q\uh \in x - B\trans \partial \reg( B\uh) \; .
\label{eq:grad}
\eeq
To make use of this observation, we introduce the affine transformation $A:\R^m \to \R^m$ defined, for fixed $x\in\R^d$, $\lambda>0$, at $z\in\R^m$ by
\beq
Az := (I-\lambda BQ^{-1}B\trans) z + BQ^{-1}x
\label{eq:A}
\eeq
and the nonlinear operator $H:\R^m\to \R^m$ 
\beq
H := \left(I-\prox_{\frac{\reg}{\lam}} \right)\circ A \;.
\label{eq:H}
\eeq
The next theorem from \cite{andy-tech} is a natural extension 
of an observation in \cite{MSX}, which only applies to the case $Q=I$. 

\begin{theorem}
If $\reg$ is a convex function on $\R^m$, $B\in\R^{m \times d}$, $x\in\R^d$, $\lambda$
is a positive number, the operator $H$ is defined as in \eqref{eq:H}, and $\uh$ is the minimizer of \eqref{eq:quad} then
\beq
\uh = Q^{-1}(x - \lambda B\trans v)
\label{eq:fixed}
\eeq
if and only if $v\in\R^m$ is a fixed point of $H$.
\label{thm:fixed}
\end{theorem}

This theorem provides us with a practical tool to solve problem \eqref{eq:quad} numerically by using Picard iteration relative to the nonlinear mapping $H$. Under an additional hypothesis on the matrix $B Q^{-1} Q\trans$, the mapping $H$ is non-expansive, see \cite{andy-tech}. Therefore, Opial's Theorem \cite{zalinescu} allows us to conclude that the Picard iterate converges to the solution of \eqref{eq:quad}, see \cite{andy-tech,MSX} for a discussion of this issue.
Furthermore, under additional hypotheses the mapping $H$ is a contraction. In that case, the Picard iterate converges linearly.

We may extend the range of applicability of our observations and provide a fixed point proximal-gradient method for solving problem \eqref{eq:opt} when the regularizer has the form \eqref{eq:comp-reg} and the error $f$ is a {\em strongly smooth} convex function, that is, the gradient of $f$, denote by $\nabla f$, is Lipschitz continuous with constant $L$. So far, the convergence of this extension has yet to be analyzed. The idea behind proximal-gradient methods, see \cite{combettes,Nesterov07,tseng10} and references therein, is to update the current estimate of the solution $x_t$ using the proximity operator of $g$ and the gradient of $f$. This is equivalent to replacing $f$ with its linear approximation around a point which is a function of the previous iterates of the algorithm. The simplest instance of this iterative algorithm is given in Algorithm 1 \ref{alg:prox}. Extensions to acceleration schemes are described in \cite{andy-tech}.

\begin{algorithm}
\caption{Proximal-gradient \& fixed point algorithm.}
\begin{algorithmic}
\STATE $x_1 \leftarrow 0$
\FOR {t=1,2,\dots}
\STATE Compute $x_{t+1} \leftarrow 
\prox_{\frac{\reg}{L}\circ B}\left(x_t - \frac{1}{L}\nabla f(x_t)\right)$
\\\qquad by the Picard process. 
\ENDFOR 
\end{algorithmic}
\label{alg:prox}
\end{algorithm}

\subsection{Connection to the forward-backward algorithm}
\label{sec:ista}

In this section, we consider the special case $Q=I$ and 
interpret the Picard iteration of $H$ in terms of a {\em forward-backward algorithm} in the dual,  for a discussion of the forward-backward algorithm, see for example \cite{combettes} 

The Picard iteration is defined as
\beq
v_{t+1} \leftarrow
(I - \prox_{\frac{\reg}{\lambda}}) ((I-\lam BB\trans)v_t + Bx)
\label{eq:picard}
\eeq

We first recall the Moreau decomposition, see, for example, \cite{combettes} and references therein, which 
relates the proximity operators of a lower semicontinuous convex
function $\varphi: \R^m \to \R \cup \{+\infty\}$ and its conjugate,
\beq
I = \prox_{\varphi} +\, \prox_{\varphi^*} \;.
\label{eq:moreau}
\eeq
Using equation \eqref{eq:moreau}, the iterative step \eqref{eq:picard} becomes  
\beq
v_{t+1} \leftarrow
\prox_{\left(\frac{\omega}{\lam}\right)^*} \,
( v_t-(\lam BB\trans v_t -  Bx))
\eeq
which is a forward-backward method. We can further simplify this iteration by introducing the vector 
$z_t := \lam v_t$ and obtaining the iterative algorithm
\beq
z_{t+1} \leftarrow
\lam \prox_{\left(\frac{\omega}{\lam}\right)^*} \,
\left( \frac{1}{\lam} z_t-( BB\trans z_t -   Bx) \right) \;.
\eeq
Using the readily verified formulas 
\beq
\frac{1}{\lam} \prox_{\lam g} \circ \lam I = \prox_{\frac{1}{\lam} g\circ\lam I}
\eeq
and
\beq
\left(\frac{\omega}{\lam}\right)^* = \frac{1}{\lam}\omega^*\circ \lam I
\eeq
see, for example, \cite{borwein}, we obtain the equivalent forward-backward iteration
\beq
z_{t+1} \leftarrow
\prox_{\lam \omega^*} (z_t-(\lam BB\trans z_t -  \lam Bx)) \;.
\label{eq:fwd_bwd}
\eeq
This method is a forward-backward method 
of the type considered in \cite[Alg. 10.3]{pesquet} and solves the minimization problem
\beq
\min\left\{ \frac{1}{2} \| B\trans z -x\|^2 + 
 \omega^*(z) : z \in\R^m \right\}  \;.
\label{eq:dual}
\eeq
This minimization problem in turn can be viewed as the dual of the primal problem 
\beq
\min\left\{
\frac{1}{2} \|u\|^2 - \lb x, u\rb + \omega(Bu) : u \in\R^d 
\right\}
\label{eq:primal}
\eeq
by using Fenchel's duality theorem, see, for example, \cite{borwein}. Moreover, the primal and dual solutions are related
through the conditions $-B\trans {\hat z} = {\hat u}-x$ and ${\hat z} \in \partial \omega(B{\hat u})$, the first of which implies that $x-\lam B\trans {\hat v}$ 
equals the solution of the proximity problem \eqref{eq:primal},
that is, equals $\prox_{\omega\circ B}(x)$.

\section{Examples of Composite Functions}
\label{sec:3}
In this section, we provide some examples of penalty functions which have appeared in the literature 
that fall within the class of linear composite functions \eqref{eq:comp-reg}. 

We define for every $d \in \N$, $x \in \R^d$ and $J \subseteq \{1,\dots,d\}$, the 
restriction of the vector $x$ to the index set $J$ as $x_{|J} = (x_i: i \in J)$. 
Our first example considers the Group Lasso penalty function, 
which is defined as
\beq
\reg_{\rm GL}(x) = \sum_{\ell=1}^k \|x_{|J_\ell}\|_2,
\label{eq:GL}
\eeq 
where $J_\ell$ are prescribed subsets of $\{1,\dots,d\}$ (also
called the ``groups'') such that $\cup_{\ell=1}^k J_\ell = \{1,\dots,d\}$. The
standard Group Lasso penalty, see, for example, \cite{yuan}, corresponds to the
case that the collection of groups $\{J_\ell : 1 \leq \ell \leq k\}$ forms
a partition of the index set $\{1,\dots,d\}$, that is, the groups do not
overlap. In this case, the optimization problem \eqref{eq:prox} for
$\omega=\omega_{\rm GL}$ decomposes as the sum of separate problems
and the proximity operator is readily obtained by using the proximity operator of the $\ell_2$-norm 
to each group separately.  In many cases of interest, however, the groups overlap and the proximity operator
cannot be easily computed.

Note that the function \eqref{eq:GL} is of the form \eqref{eq:comp-reg}. 
We let $d_\ell = |J_\ell|$, $m =\sum_{\ell=1}^k d_\ell$ and define, for 
every $z \in \R^m$, $\reg(z) = \sum_{\ell=1}^k \|z_{\ell}\|_2$, 
where, for every $\ell=1,\dots,k$ we let $z_\ell = (z_i: \sum_{j=1}^{\ell-1} 
d_j< i \leq \sum_{j=1}^\ell d_j)$.
Moreover, we choose $B\trans = [B_1\trans,\dots,B_k\trans]$, 
where $B_\ell$ is a $d_\ell \times d$ matrix defined as
\begin{equation*}
(B_\ell)_{ij} = \left\{
\begin{array}{rl}
1 & \text{if~} j = J_\ell[i] \\
0 & \text{otherwise}
\end{array} \right. \; ,
\end{equation*}
where for every $J \subseteq \{1,\dots,d\}$ and $i \in \{1,\dots,|J|\}$, we denote by $J[i]$ 
the $i$-th largest integer in $J$. 

The second example concerns the Fused Lasso \cite{tib05}, which considers 
the penalty function $x \mapsto g(x)= \sum_{i=1}^{d-1} |x_{i}-x_{i+1}|$. 
This function falls into the class \eqref{eq:comp-reg}. Indeed,  if we choose $\reg$ to be the $\ell_1$ norm and $B$ the first order divided difference
matrix
\begin{equation}
B= \left[
\begin{array}{rrrrr}
1 & -1 & 0 & \ldots & \ldots\\
0 & 1 & - 1 & 0 & \ldots \\
\vdots & \ddots & \ddots & \ddots & \ddots\\
\end{array} \right]
\label{eq:fused}
\end{equation}
we get back $g$. The intuition behind the Fused Lasso is that it favors vectors which
do not vary much across contiguous components.  Further extensions of
this case may be obtained by choosing $B$ to be the incidence matrix
of a graph, leading to the penalty $\sum_{(i,j) \in E}^n |x_i-x_j|$. This is a setting which is relevant, for example, in online learning over graphs \cite{mark09,HP07}. 

The next example considers composition with orthogonally invariant (OI) 
norms. Specifically, we choose a symmetric gauge function $h$,
that is, a norm $h$, which is both {\em absolute} and {\em invariant under
permutations} \cite{von-neumann} and define the function
$\reg:\R^{d\times n} \rightarrow [0,\infty)$, at $X$ by the formula $\reg(X) = h(\sigma(X))$, 
where $\sigma(X) \in [0,\infty)^{r}$, $r = \min(d,n)$ is the vector
formed by the singular values of matrix $X$, in non-increasing
order. An example of OI-norm are Schatten
$p$-norms, which correspond to the case that $\reg$ is
the $\ell_p$-norm. The next proposition provides a formula for the proximity operator of 
an OI-norm.  A proof can be found in \cite{andy-tech}.

\begin{proposition}
With the above notation, it holds that 
$$
\prox_{h \circ \sigma}(X) = U {\rm diag}\left(\prox_h(\sigma(X))\right) V\trans
$$  
where $X=U{\rm diag}(\sigma(X)) V\trans$ and $U$ and $V$ are the matrices 
formed by the left and right singular vectors of $X$, respectively. 
\label{prop:prox-OI}
\end{proposition}

We can compose an OI-norm with a linear transformation $B$, this time
between two spaces of matrices, obtaining yet another subclass of
penalty functions of the form
\eqref{eq:comp-reg}.  This setting is relevant in the context of
multi-task learning. For example, in
\cite{AEP} $h$ is chosen to be
the {\em trace} or {\em nuclear} norm and a specific linear
transformation which models task relatedness is considered. Specifically, the regulariser is given by 
$g(X) = \left\|\sigma\left(X(I-\frac{1}{n}ee\trans)\right)\right\|_1$, 
where $e \in \R^d$ is the vector all of whose components are equal to one.


\section{Application to Support Vector Machines}
\label{sec:4}
In this section, we turn our attention to the important topic of support vector machines (SVMs), which are widely used in data analysis. SVMs were pioneered by the fundamental work of Vapnik \cite{Boser,CV,vapnik} and inspired one of us to begin research in machine learning \cite{EPP,PV,PPP}. For that we are all very grateful to Vladimir Vapnik for his fundamental contributions to machine learning.

First, we recall the SVM primal and dual optimization problems, \cite{vapnik}. To simplify the presentation we only consider
the linear version of SVMs. A similar treatment using feature map representations is 
straightforward and so will not be discussed here, although this in a an important extension of practical value. 
Moreover, we only consider SVMs for classification, but our approach can be applied to SVM regression and other variants of SVMs which have appeared in the literature.
 
The optimisation problem of concern here is given by 
\begin{equation}
\min \left\{ C\sum_{i=1}^m V(y_i w\trans x_i) + \frac{1}{2} \|w\|^2 : w
\in \R^d \right\}
\label{eq:svm_primal}
\end{equation}
where $V(z) = \max(0,1-z)$, $z\in\R$, is the hinge loss and $C$ is a positive
parameter balancing empirical error against margin maximization. 
We let $x_i \in \R^d,$ $i\in \{1,\dots,m\},$ be the input data and $y_i\in \{-1,+1\}$
be the class labels.

Problem \eqref{eq:svm_primal} can be viewed as a proximity operator
computation of the form \eqref{eq:quad}, with 
$Q = I$, $x=0$, $\omega(z) = C \sum_{i=1}^m V(z_i)$ 
and $B = [y_1 x_1 \dots y_m x_m]\trans$.
The proximity operator of the hinge loss is separable across the
coordinates and simple to compute. In fact, for any $\zeta\in\R$ and $\mu > 0$ it is given by the formula
 \beq
 \prox_{\mu V}(\zeta) =
\min(\zeta+\mu,\max(\zeta,1)).
 \label{eq:svm_prox}
 \eeq
Hence, we can solve problem \eqref{eq:svm_primal} by Picard iteration, namely
\beq
v_{t+1} \leftarrow
\left(I-\prox_{\frac{\omega}{\lam}}\right)
\left((I-\lam BB\trans)v_t\right)
\label{eq:picard_svm}
\eeq
with $\lam$ satisfying
$0<\lam < \frac{2}{\lam_{\rm max}(BB\trans)}$, which ensures that the nonlinear mapping is strictly contractive. 
Note that $v_t \in \R^m$ and that this iterative scheme may be interpreted as acting 
on the SVM dual, see Section \ref{sec:ista}. In fact, there is a
simple relation to the support vector coefficients given by the equation 
$v = \frac{1}{\lam}\alpha$. 
Consequently, this algorithmic approach is well suited when the sample size $m$ is
small compared to the dimensionality $d$.
An estimate of the primal solution, if required, can be obtained by using the formula $w = -\lam B\trans v$.
Also, when $d<m$ the last equation, relating $w$ and $v$, cannot be inverted. Hence, \eqref{eq:picard_svm} is not useful in this case.


Recall that the dual problem of \eqref{eq:svm_primal} is given \cite{vapnik}
\begin{equation}
\min \left\{\frac{1}{2} \|B\trans \alpha\|^2
- 1\trans \alpha: \; \alpha \in [0,C]^m \right\}.
\label{eq:svm_dual}
\end{equation}
This problem can be seen as the computation of a generalized proximity operator
of the type \eqref{eq:quad}. To explain what we have in mind we use the notation $\odot$ as the elementwise product between matrices of the same size (Schur product) and introduce the kernel matrix $K = [x_1 \dots x_m]\trans [x_1 \dots x_m]$. 

Using this terminology, we conclude that problem \eqref{eq:svm_dual} is of the form \eqref{eq:quad} with $Q = K \odot yy\trans$, $x= \mathbf{1}$ (the vector of all ones), $B=I$ and $\omega = \omega_C$, where $\omega_C(\alpha) = 0$ if $\alpha \in [0,C]^m$ and $\omega_C(\alpha) = +\infty$ otherwise. Furthermore, the proximity operator for $\omega$ is given by the projection on the set $[0,C]^m$, that is $\prox_{\omega_C}(\alpha) = \min(C,\max(0, \alpha))$. These observations yield the Picard iteration
\beq
v_{t+1} \leftarrow 
\left(I-\prox_{\omega_C} \right) 
\left( (I-\lambda (K^{-1} \odot yy\trans)) v_t + (K^{-1} \odot yy\trans)\mathbf{1} \right) 
\label{eq:picard_svm_dual}
\eeq
with $0< \lam < 2\,\lam_{\rm min}(K)$. This iterative scheme requires that the
kernel matrix $K$ is invertible, which is frequently the case, for
example, in the case of Gaussian kernels.
Another requirement is that either $K^{-1}$ has to be
precomputed or a linear system involving $K$ has to be solved at every
iteration, which limits the scalability of this scheme to very large samples.
In contrast, the iteration \eqref{eq:picard_svm} can always be
applied, even when $K$ is not invertible. In fact, when $K$, and equivalently $BB\trans$, is invertible then
both iterative methods \eqref{eq:picard_svm}, \eqref{eq:picard_svm_dual} converge linearly at a rate which depends
on the condition number of $K$, see \cite{andy-tech,MSX}. 

Recall that algorithm \eqref{eq:picard_svm} is equivalent to a
forward-backward method in the dual, see Section \ref{sec:ista}.
Thus, an accelerated variant akin to Nesterov's optimal method and
FISTA \cite{fista} could also be used. However, in the case of an invertible kernel matrix, both
versions converge linearly \cite{Nesterov07} and hence it is not
clear whether there is any practical advantage from the Nesterov update. 
Furthermore, algorithm \eqref{eq:picard_svm_dual} could also be modified in a
similar way.

On the other hand, if $m>d$, we would directly attempt to solve the primal problem. 
In this case, the Nesterov smoothing method can be employed, \cite{nesterov2005smooth}.  
An advantage of such a method is that it only stores $O(d)$ variables, even though it needs
$O(md)$ computations per iteration. The method described above, based on Picard iteration, requires $\min(O(md), O(m^2))$ cost per iteration and stores $O(m)$ variables.

Let us finally remark that iterative methods similar to
\eqref{eq:picard_svm} or \eqref{eq:picard_svm_dual} can be applied to $\ell_2$
regularization problems, other than SVMs, provided that the proximity operator of the corresponding loss function is available. Common choices for the loss 
function, other than the hinge loss, are the logistic and square 
loss functions leading to logistic regression and least squares
regression, respectively. In particular, in these two cases, the primal objective
\eqref{eq:svm_primal} is both smooth and strongly convex
and hence a linearly convergent gradient descent or accelerated
gradient descent method can be used \cite{nesterov_book}, 
regardless of the conditioning of the kernel matrix.


\section{Conclusion}
\label{sec:5}
We presented a general approach to solve a class of nonsmooth optimization problems, whose objective function is given by the sum of a smooth term and a nonsmooth term which is obtained by linear function composition. The prototypical example covered by this setting is a linear regression regularization method, in which the smooth term is an error term and the nonsmooth term is a regularizer which favors certain desired parameter vectors. An important feature of our approach is that it can deal with a rich class of
regularizers and, as shown numerically in \cite{andy-tech}, is competitive with the state of the art methods. 
Using these ideas, we also provided a fixed-point scheme to solve support vector machines. Although numerical experiments have yet to be done, we believe this method is simple enough to deserve attention by practitioners. 

We believe that the method presented here should be throughly investigated both in terms of convergence analysis, where ideas presented in \cite{villa} may be valuable, and numerical performance with other methods, such as alternate direction of multipliers, see, for example, \cite{Boyd}, block coordinate descent, alternate minimization and others. Finally, there are several other machine learning problems where ideas presented here apply. For example, in that regard we mention multiple kernel learning, see for example, \cite{MP07,mkl,mkl2,suzuki} and references therein, some structured sparsity regularizers \cite{MauPon,MMP} and multi-task learning, see, for example \cite{AEP,CCG,EPT}. We leave these tantalizing issues for future investigation.

\subsubsection*{Acknowledgements}
Part of this work was supported by EPSRC Grant
EP/H027203/1, Royal Society International Joint
Project Grant 2012/R2 and by the European Union Seventh Framework Programme (FP7 2007-2013) under grant agreement No. 246556.

\end{document}